%% file: main.tex
\documentclass{article}

\newif\ifanonymous
\anonymousfalse

\ifanonymous
  \usepackage{neurips_2026}
\else
  \usepackage[preprint]{neurips_2026}
\fi

\usepackage[utf8]{inputenc}
\usepackage[T1]{fontenc}
\usepackage{hyperref}
\usepackage{url}
\usepackage{booktabs}
\usepackage{graphicx}
\usepackage{amsmath}
\usepackage{amssymb}
\usepackage{amsfonts}
\usepackage{nicefrac}
\usepackage{microtype}
\usepackage{xcolor}

\usepackage{multirow}
\usepackage{subcaption}
\usepackage{tabularx}
\usepackage{makecell}
\usepackage{adjustbox}
\usepackage{float}
\newcolumntype{Y}{>{\centering\arraybackslash}X}
\usepackage{tikz}
\usetikzlibrary{positioning,fit,arrows.meta}

\title{World Models Under Asynchronous Sensor Observations}

\author{%
  Akash Anand$^{1*}$
  \qquad
  Abhay Anand$^{2*}$
  \qquad
  Yash Vishe$^{2*}$ \\[-1pt]
  $^{1}$University of California, Davis
  \qquad
  $^{2}$University of California, San Diego \\[-1pt]
  $^{*}$Equal contribution
}

\begin{document}

\maketitle

\begin{abstract}
% Learned world models assume that every observation arrives on the same clock. Real sensors do not oblige: an IMU reports hundreds of times a second, a camera tens, GPS once. At any instant, most channels are showing a value that is already out of date. The natural fix is to hold the last reading and tell the model about the schedule itself --- how stale each channel is, and how long until it refreshes. We argue that only half of this fix can work, and we explain why. A schedule says when a model will learn something, not what is true. So long as sensors do not act on the system they measure, the time until the next reading cannot improve prediction at any horizon; staleness, which describes what has already been observed, is a different matter. This predicts that refresh timing pays only when a refresh changes the world rather than merely reporting on it, and we test that ordering in three settings chosen to vary exactly that. In open-loop prediction and under a planner, timing information does nothing, and our measurements bound the effect rather than merely failing to find one. Where a refresh latches a held actuator, the same models gain a great deal, and gain more as the schedule grows more irregular. Telling a world model when its sensors will report is wasted effort; telling it when its actuators will update is not.

Learned world models typically assume that observations arrive synchronously, an abstraction inherited from simulators that return a complete state vector at each environment step. Physical sensing instead operates at heterogeneous rates, leaving most observation channels stale at any given instant. Interpolating stale channels introduces measurements that were never observed, while downsampling to the slowest sensor discards valid measurements. A natural alternative is to zero-order-hold the most recent reading and provide the known sampling schedule to the model through two features, staleness and time-to-refresh. We test this prediction using transformer world models across three regimes of increasing causal coupling: open-loop rollouts in continuous-control locomotion, closed-loop model-predictive planning in which each learned model serves as the planner dynamics, and a linear latched-actuator system in which refresh events apply a zero-order-held command to the plant. Our findings show that the effectiveness of time-to-refresh depends on the causal role of the sampling schedule, specifically when refresh events affect the system rather than merely report its state. These results establish when sampling schedules provide useful information for predictive world models operating under asynchronous physical observations.
\end{abstract}

\input{sections/introduction}
\input{sections/method}
\input{sections/experiments_results}
\input{sections/related_work}
\input{sections/conclusion}

\clearpage
\bibliographystyle{plainnat}
\bibliography{references}

\appendix
\input{sections/appendix}

% ---------------------------------------------------------------------
% Mandatory. Follows the references and the appendix. Does not count
% toward the page limit. Papers without it are desk rejected.
% ---------------------------------------------------------------------

\end{document}

%% file: sections/introduction.tex
\section{Introduction}

World models are predominantly developed in simulated environments \citep{hafner2020dreamer,hafner2023dreamerv3,micheli2023iris,alonso2024diamond,hafner2019planet,hansen2022tdmpc}, where observations are available synchronously at every timestep. This assumption is natural in simulation, where the environment can return the complete observation whenever it is queried. Physical systems do not provide the same abstraction. Sensors operate at heterogeneous rates, leaving parts of the observation stale between successive measurements. As a result, world models deployed in physical systems must operate on observations that are neither synchronized nor uniformly current. Existing approaches to reconciling these rates come with trade-offs. Interpolation constructs measurements that were never observed and may require future samples unavailable at inference time, while downsampling to the slowest sensor discards measurements from faster channels. A natural alternative is to zero-order-hold each channel at its most recent measurement and provide its sampling schedule to the model. Since sensor refresh schedules are often known in advance, two features can be derived directly from the schedule: staleness, which measures the time since the previous refresh, and time-to-refresh (ttr), which measures the time until the next. While both describe the sampling schedule, they provide fundamentally different information.

This work studies whether these two forms of schedule information contribute equally to world-model prediction. To isolate their contributions, we construct a ladder of transformer world models that progressively receive more information about the sampling schedule. A-ZOH observes only the most recent measurement from each channel, F-blind additionally receives staleness, and F receives both staleness and time-to-refresh. We evaluate these models across three settings with increasing causal coupling between the sampling schedule and the underlying system. We first study open-loop rollouts in three MuJoCo locomotion environments \citep{todorov2012mujoco}, varying the sampling schedule from periodic refreshes to $\pm50\%$ jitter. We evaluate these rollouts under both a cut-off protocol, where no new sensor observations are provided during prediction, and a sensor-faithful protocol, where observations are updated according to the underlying refresh schedule. We then place the learned models inside a fixed model-predictive controller, where model predictions influence actions and future states. Finally, we construct paired linear systems that differ only in the role of the refresh event. In the reveal setting, refreshes only provide new observations, while in the causal setting, refresh timing determines when a held actuator command is applied to the system. This final setting provides a controlled case where time-to-refresh should contain information about the system's future evolution.

Our experiments show a consistent separation between these settings. In open-loop prediction, providing time-to-refresh produces no measurable improvement, with a pooled sensor-faithful effect of $-0.0005 \pm 0.0066$. The effect remains within noise under model-predictive control, where differences due to schedule information are substantially smaller than ordinary world-model error. In contrast, when refresh events control when an actuator is updated, time-to-refresh allows the model to recover $7.31$ of the $9.14$ divergence between two otherwise identical systems ($80\%$), against $3.02$ without it ($33\%$), and the effect increases as the refresh schedule becomes less predictable. These results establish a causal criterion for when forward-looking schedule information is useful in world models and provide a basis for designing world models that operate beyond the synchronized observation setting assumed by standard simulation environments.

%% file: sections/method.tex
\section{Method}

\begin{figure}[t]
    \centering
    \includegraphics[width=\linewidth]{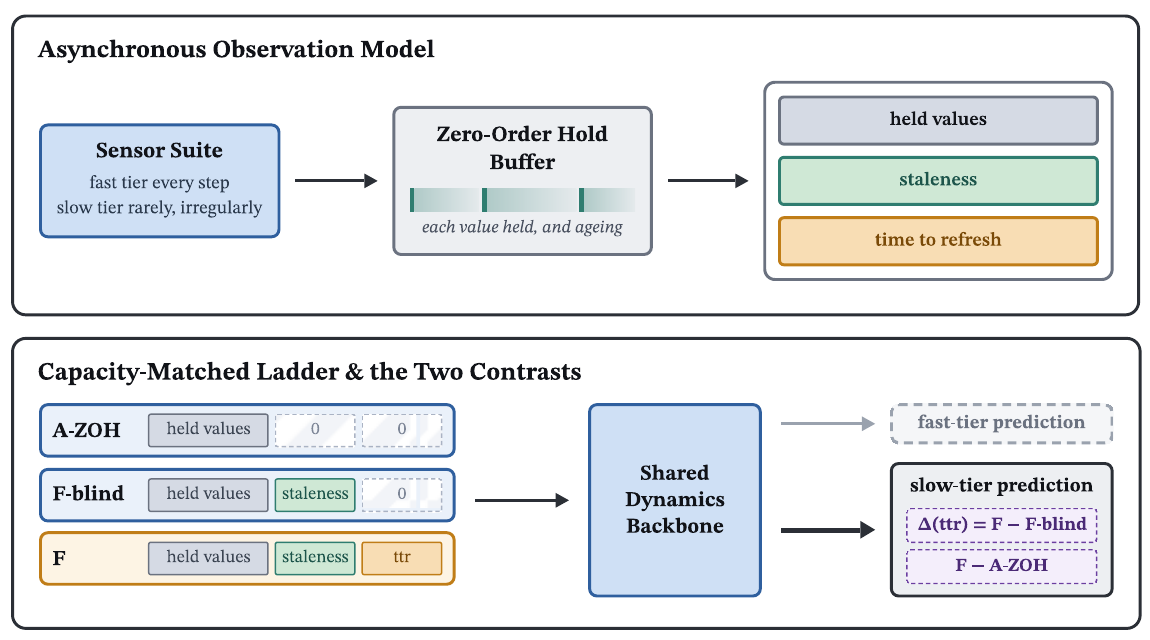}
    \caption{Overview of the asynchronous observation model and the model ladder.
    Each sensor channel is held at its most recent measurement between refreshes and accompanied by
    staleness and time-to-refresh. The rungs share the same dynamics architecture; F-blind and F
    additionally share an input width, with ttr zeroed for F-blind. The primary contrast,
    $\text{F}-\text{F-blind}$, isolates the contribution of time-to-refresh.}
    \label{fig:overview}
\end{figure}

World models deployed with heterogeneous sensors observe a mixture of fresh and stale
measurements rather than a complete state updated on one clock. We represent this setting
explicitly. The central distinction is between information about the observation history and
information about the future observation schedule: the first can constrain the current state,
whereas the second need not constrain how that state will evolve.

\subsection{Asynchronous observations and schedules}
\label{sec:async}

Let $s_t\in\mathbb{R}^d$ be the underlying state at time $t$ and $a_t$ the action. Each
observation channel $i$ refreshes on its own schedule. Between refreshes, its latest measurement is
zero-order-held rather than interpolated. Writing
$\tau_i(t)$ for the most recent refresh and
$\nu_i(t)$ for the next scheduled refresh, the model receives
\begin{equation}
  o_t^{(i)} = s_{\tau_i(t)}^{(i)}, \qquad
  \delta_t^{(i)} = t-\tau_i(t), \qquad
  \rho_t^{(i)} = \nu_i(t)-t .
\end{equation}
Here $o_t^{(i)}$ is the held observation, $\delta_t^{(i)}$ its \emph{staleness}, and
$\rho_t^{(i)}$ its \emph{time-to-refresh} (ttr). Both timing features are normalized by the
channel's nominal period. Staleness looks backward and records how long a value has been
propagated; ttr looks forward and records when that value will next be corrected. This is not
merely a directional difference. Staleness describes what has already been observed and therefore
changes the meaning of a held value. Ttr says when the model will learn something next, but does
not by itself say what will be learned.

Let $S_{\le t}$ and $S_{>t}$ denote the observed and future refresh schedules, with
$o_{\le t}$ the observation history and $a$ the action sequence. If future sensing is exogenous to
the system dynamics,
\begin{equation}
  S_{>t} \perp s_{t+k} \mid (o_{\le t}, S_{\le t}, a),
  \label{eq:exogeneity}
\end{equation}
then conditioning on ttr leaves $\mathbb{E}[s_{t+k}\mid\cdot]$ unchanged and cannot reduce the
Bayes risk of point prediction at any horizon. Staleness is not subject to this argument because
it belongs to the conditioning history. Thus the split is structural: one feature describes what
is known, while the other describes when uncertainty will next be resolved.

The claim is deliberately conditional. It applies when the refresh process is fixed independently
of the plant and the target is a point prediction of future state. It does not cover event-triggered
sensors whose sampling decision depends on the state, active sensing policies that choose what to
measure, or schedules coupled to a shared controller or communication bus. In those cases the
schedule can itself reveal state or participate in the transition. Nor does
Equation~\eqref{eq:exogeneity} imply that ttr is useless for every downstream task: knowing when a
measurement will arrive may still matter for uncertainty calibration, computation scheduling, or
measurement selection. Our question is narrower and testable: after conditioning on the available
history and actions, does ttr improve the state trajectory predicted by an action-conditioned world
model? This distinction prevents benefits for resource allocation from being conflated with
information about the state itself.

Channels are assigned to fast, medium, and slow tiers by sensor semantics rather than by measured
volatility. Slow channels represent quantities that are physically expensive to measure or are
naturally available only at a lower rate, such as root-body velocity estimated from multiple
sensors. Under the
\texttt{moderate} setting used throughout, fast, medium, and slow channels have nominal periods
$1$, $5$, and $10$ steps.

We vary the predictability of the refresh times while holding these mean periods fixed.
\textbf{Periodic} schedules refresh exactly every $P$ steps, so the next refresh is recoverable
from phase. Under \textbf{$\pm10\%$} jitter, the slow tier varies by one step. Under
\textbf{$\pm50\%$} jitter, intervals are drawn as integers from $[P/2,3P/2]$, making the next
refresh unavailable from the step count alone. Jitter applies only to channels with $P>1$; the
fast tier refreshes at every step and is never jittered. Jitter therefore acts as a dose-response
test rather than only a robustness perturbation. If ttr helps because of the information it
carries, its advantage should grow from periodic to $\pm50\%$ jitter; a flat or shrinking effect
indicates that the feature is not being used for its content.

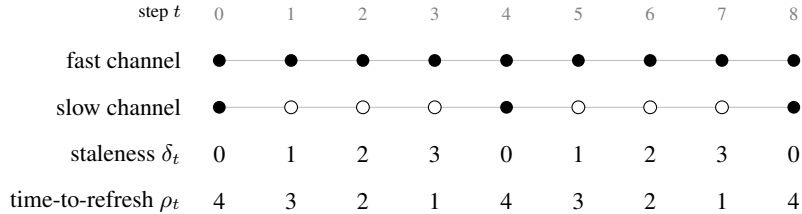
\begin{figure}[H]
\centering
\begin{tikzpicture}[font=\small, x=0.95cm, y=0.62cm]
  \node[anchor=east] at (-0.4,4) {\scriptsize step $t$};
  \foreach \t in {0,...,8} { \node[gray] at (\t,4) {\scriptsize \t}; }

  \node[anchor=east] at (-0.4,3) {fast channel};
  \draw[gray!50] (0,3) -- (8,3);
  \foreach \t in {0,...,8} { \fill (\t,3) circle (2.4pt); }

  \node[anchor=east] at (-0.4,2) {slow channel};
  \draw[gray!50] (0,2) -- (8,2);
  \foreach \t in {1,2,3,5,6,7} { \draw[fill=white] (\t,2) circle (2.4pt); }
  \foreach \t in {0,4,8} { \fill (\t,2) circle (2.4pt); }

  \node[anchor=east] at (-0.4,1) {staleness $\delta_t$};
  \foreach \t/\v in {0/0,1/1,2/2,3/3,4/0,5/1,6/2,7/3,8/0} { \node at (\t,1) {\v}; }

  \node[anchor=east] at (-0.4,0) {time-to-refresh $\rho_t$};
  \foreach \t/\v in {0/4,1/3,2/2,3/1,4/4,5/3,6/2,7/1,8/4} { \node at (\t,0) {\v}; }
\end{tikzpicture}
\caption{Schedule features for a slow channel with period 4, shortened from the period 10 used in
the experiments. Filled circles mark refreshes and open circles held values. Staleness is
determined by past observations, whereas time-to-refresh describes the future schedule.}
\label{fig:schedule}
\end{figure}

\subsection{Predicted ordering across three settings}
\label{sec:ordering}

Equation~\eqref{eq:exogeneity} predicts an ordering by the causal role of a refresh, not merely a
null result in one benchmark. We test three settings while holding the model ladder and its primary
contrast fixed. \textbf{Act 1} is open-loop prediction, where a refresh only reveals state and ttr
should be inert. \textbf{Act 2} places the model inside a planner. Observations now influence
actions and actions influence state, creating an indirect and weak schedule-to-state path; a small
effect is possible, but direct model error may dominate it. \textbf{Act 3} makes the refresh itself
causal by latching a held actuator command. Ttr then identifies an event in the dynamics and should
be useful precisely when jitter makes that event otherwise unpredictable.

The third setting is a positive control for the first two. Without it, ``no effect found'' cannot
be separated from an insensitive model or metric. Recovering a ttr advantage under refresh-gated
actuation---and recovering a larger advantage as jitter increases---would show that the same
experimental instrument can detect the predicted effect when the schedule actually enters the
dynamics.

The three acts vary two properties that should be read separately. The first is \emph{causal
strength}: whether a refresh merely reports the system, affects it indirectly through planning, or
directly changes the applied control. The second is \emph{inferability}: whether the next refresh
can already be recovered from periodic phase or is genuinely new information under jitter. The
theory predicts no advantage when causal strength is absent, and no incremental advantage when the
timing is redundant. A ttr effect should therefore appear only where the refresh is causal and the
timing is not already inferable from phase. This
crossed logic is stronger than asking whether F wins in any single cell, because it specifies the
direction in which the contrast must move before training.

\subsection{Capacity-matched model ladder}
\label{sec:ladder}

We train three world models that differ only in the schedule information carried by their inputs.
We call them \emph{rungs}, since each adds one piece of schedule information to the one below:
\begin{equation}
  \text{A-ZOH}:\;o_t, \qquad
  \text{F-blind}:\;(o_t,\delta_t), \qquad
  \text{F}:\;(o_t,\delta_t,\rho_t).
\end{equation}
A-ZOH receives only the held observations. F-blind adds the age of each observation, allowing it
to distinguish a recent measurement from one propagated for several steps. F additionally
receives the next refresh time.

F-blind and F retain the same input width and parameter count: ttr is zeroed for F-blind rather
than removed, so the two differ in information alone. Our primary contrast is
\begin{equation}
  \Delta(\mathrm{ttr}) = \text{F}-\text{F-blind},
\end{equation}
where a positive value indicates an advantage from ttr.

This matching is essential to the interpretation. Removing the ttr column would also change
the first projection and parameter count, leaving model capacity as an alternative explanation for
any gain. Zeroing instead keeps the architecture, optimiser, data, and input width fixed while
changing only the information carried by that column. The adjacent contrast
$\text{F}-\text{F-blind}$ therefore isolates forward-looking timing.

The current held observation, action, and schedule features are projected into a common token,
processed by a causal Transformer, and mapped through a linear head to the next observation.

\subsection{Data collection and training}
\label{sec:data}

The prediction and control studies use Ant-v4, HalfCheetah-v4, and Walker2d-v4
\citep{todorov2012mujoco,towers2024gymnasium}, with root linear and angular velocity assigned to
the slow tier. Trajectories come from soft actor-critic (SAC) policies \citep{haarnoja2018sac}
trained for 300{,}000 steps, keeping the data on
the state-action distribution relevant to control. The grid crosses three rungs, three
environments, three schedules, and three seeds, for 81 runs. Episodes are split 80/10/10 into
train, validation, and test sets; channel statistics come from training only, and schedules are
reconstructed deterministically from the seed and episode index.

For the refresh-gated study, each variant, schedule, and seed uses its own schedule realization;
otherwise ttr could point to steps where no latch occurred.

Every rung uses the same roughly 0.8M-parameter GPT-2-style causal Transformer and is trained for
15 epochs with AdamW. Architecture and optimization are identical across rungs and environments;
dataset sizes and the complete configuration are reported in Appendix~\ref{app:setup}.

\subsection{Rollout evaluation}
\label{sec:eval}

Models are trained to predict one step ahead and evaluated over multi-step rollouts, as is
standard; Appendix~\ref{app:rollout} reports a fine-tuning study confirming that this choice does
not drive the result. After the
rollout begins, the model consumes its own predictions over horizons $H\in\{5,15,50\}$.
Metrics are computed separately at each rollout step and for each sensor tier. The primary
prediction measure is the slow-tier value at the final evaluated step, $H{=}50$.

Raw mean-squared error can reward copying a slowly changing stale value. We therefore report skill
relative to the zero-order-hold baseline,
\begin{equation}
  \mathrm{Skill}_{\mathrm{ZOH}}
  =
  1-\frac{\mathrm{MSE}_{\mathrm{model}}}{\mathrm{MSE}_{\mathrm{ZOH}}},
\end{equation}
where positive values indicate that the model improves on holding fixed the last real sensor
measurement available at the start of the rollout.

This matters most on the slow tier, where even a more informative model can look worse if it
attempts to predict motion between refreshes. ZOH skill makes holding the zero point and asks
whether learned dynamics add value beyond it. Rung means are reported together with paired,
within-cell ttr contrasts, so schedule information is compared under the same environment,
schedule realization, and seed.

Every prediction is evaluated under two buffer-update protocols. Under \textbf{cut-off}, a
scheduled refresh writes the model's prediction into the held buffer, so no new external
observation enters during the rollout. This models open-loop planning, but ttr announces a
correction that never arrives, so a cut-off null is not sufficient by itself. Under
\textbf{sensor-faithful}, a refresh writes the real sensor measurement, matching how a deployed
sensor corrects the held state. Predictions remain causal: a target is scored before its value can
enter the buffer, and observations appear only when their channels are scheduled to refresh. Under
both protocols, however, the ZOH baseline remains fixed at the last context measurement and is not
refreshed during the rollout. Sensor-faithful evaluation therefore scores a refresh-fed model
against a held baseline, which can drive absolute skill toward its ceiling. We interpret the two
protocols together rather than as interchangeable replications.

The two answer complementary questions: cut-off asks whether the model can carry its own belief
forward when external sensing disappears, while sensor-faithful supplies the deployment semantics
ttr promises, in which a real observation does arrive when the countdown reaches zero. Agreement
across them is therefore more informative than either alone.

\subsection{Closed-loop control}
\label{sec:control}

Each learned model is also placed inside the same cross-entropy-method controller with access to
the true reward, leaving the dynamics model as the only manipulated component. This creates an
indirect sensing-to-state path through the chosen action, although refreshes still do not change
the dynamics. We report mean episode return; controller and evaluation settings are given in
Appendix~\ref{app:setup}.

\subsection{Refresh-gated positive control}
\label{sec:toy}

To test a refresh that directly changes the dynamics, we construct two environments sharing the
linear dynamics
\begin{equation}
  v_{t+1}=\alpha v_t+u_{t+1}, \qquad
  p_{t+1}=p_t+\Delta t\,v_{t+1}, \qquad
  \alpha=0.90, \quad \Delta t=0.10.
\end{equation}
The applied control $u$ is fast, velocity and position are slow, and commands are sampled i.i.d.\
from a uniform distribution. The velocity decay $\alpha$ is distinct from the time-to-refresh
$\rho_t$ of \S\ref{sec:async}.

The environments differ only in how the command $a$ becomes the applied control $u$:
\begin{equation}
\begin{aligned}
  \textbf{reveal}:&\quad u_{t+1}=a_{t+1},\\[2pt]
  \textbf{causal}:&\quad u_{t+1}=
  \begin{cases}
    a_{t+1}, & \text{if the gating channel refreshes at } t+1,\\
    u_t,     & \text{otherwise.}
  \end{cases}
\end{aligned}
\end{equation}
A single representative slow channel, the velocity channel, gates the latch for the whole control
vector.
In \textbf{reveal}, commands apply immediately and refreshes only reveal state. In
\textbf{causal}, commands latch only on refresh, so the schedule enters the dynamics. Both use the
same model ladder. The design predicts no ttr effect in reveal or under a periodic causal
schedule, but a positive effect under causal jitter, where ttr uniquely identifies the latch. The
sensor-faithful protocol yields a fourth prediction: the arriving measurement already reveals the
latch, so the advantage should disappear.

\paragraph{Paired-world probe.}
Because average ZOH skill cannot distinguish correct latch timing from hedging, we compare matched
worlds that share an initial state and command stream and differ only in when the next slow
channel refreshes, at step 37 or 44, both valid under $\pm50\%$ jitter; until step 37 they agree
in every observable, so ttr alone separates them. We report the peak separation of the two
predicted velocity trajectories in place of ZOH skill. Each variant's target is set by the worlds,
not by any model, as the peak separation of the two \emph{true} trajectories: $9.14$ under causal
dynamics, where the worlds latch different commands and the decay carries each mistimed command
forward, and exactly zero under reveal, where the worlds are bit-identical, so any separation
there is spurious and reveal controls for reactivity to the changed schedule alone.

%% file: sections/experiments_results.tex
\section{Experiments and Results}
\label{sec:results}

Tables~\ref{tab:prediction_control} and~\ref{tab:toy_probe} report the three model runs
together with the paired contrast
$\Delta(\mathrm{ttr}) = \text{F} - \text{F-blind}$. Positive values indicate that adding
time-to-refresh improves performance. We call one row of a table a \emph{cell}: a single
combination of environment, schedule and evaluation protocol. Prediction cells are averaged over
three training seeds; control cells are averaged over evaluation seeds --- six for Ant and three
for HalfCheetah and Walker2d. Intervals are computed from paired differences on the corresponding
repeat axis.

\input{sections/table_main}

\subsection{Act 1: exogenous sensing}

% In open-loop prediction, adding ttr does not produce a consistent improvement. Pooled
% $\Delta(\mathrm{ttr})$ is $+0.0310 \pm 0.0490$ under cut-off rollout and $-0.0005 \pm 0.0066$
% under sensor-faithful rollout. Individual effects are small and change sign across environments:
% of the eighteen prediction cells, one has an interval excluding zero, and it is negative. Nor do
% the effects grow as the schedule becomes less predictable, which is what the hypothesis requires,
% since only under jitter does ttr carry timing the step count does not already supply; the
% intermediate $\pm$10\% rows show no advantage either. Nor does the contrast widen as the rollout lengthens
% (Table~\ref{tab:horizon}), and fine-tuning the same checkpoints with the loss computed on the
% rollout itself leaves it at $-0.0043 \pm 0.0051$ (Appendix~\ref{app:rollout}). The tight
% sensor-faithful interval bounds any useful effect to about $\pm0.007$, rather than
% merely failing to reject a null. When refreshes only reveal the state, knowing their future
% timing adds little to prediction.

In open-loop prediction, adding ttr does not produce a consistent improvement. Pooled $\Delta(\mathrm{ttr})$ is $+0.0310 \pm 0.0490$ under cut-off rollout and $-0.0005 \pm 0.0066$ under sensor-faithful rollout. Individual effects are small and change sign across environments: of the eighteen prediction cells, one has an interval excluding zero, and it is negative. The effects also do not grow as the schedule becomes less predictable, which is what the hypothesis requires, since only under jitter does ttr carry timing the step count does not already supply; the intermediate ±10\% rows show no advantage either. The contrast likewise does not widen as the rollout lengthens (Table~\ref{tab:horizon}), and fine-tuning the same checkpoints with the loss computed on the rollout itself leaves it at $-0.0043 \pm 0.0051$ (Appendix~\ref{app:rollout}). The tight sensor-faithful interval bounds any useful effect to about $\pm0.007$, rather than merely failing to reject a null. When refreshes only reveal the state, knowing their future timing adds little to prediction.

\subsection{Act 2: weak coupling through control}

Control gives no evidence that ttr improves planning. Returns on Ant run from $254.5$ to
$408.6$ across its six cells, against $38.3$ to $167.5$ on HalfCheetah and $18.3$ to $216.2$ on
Walker2d, so Ant is the only environment where the planner performs well enough for its rungs to
be compared on equal footing; elsewhere the contrasts carry intervals as wide as $\pm 390$ and
separate one poorly performing planner from another rather than one level of schedule information
from another. On Ant, $\Delta(\mathrm{ttr})$ is $+31 \pm 102$ at periodic, an interval containing
zero, and $-73 \pm 44$ under $\pm 50\%$ jitter. The latter is the only interval in the control
block that excludes zero, and it does so in the negative direction: F planned worse than F-blind.
No cell shows ttr helping, and control performance is limited more by the quality of the learned
dynamics than by access to ttr.

\subsection{Act 3: refresh-gated actuation}

The positive control shows the setting in which ttr does help. Under causal actuation with
$\pm 50\%$ jitter, F recovers $7.31$ of the target $9.14$ separation, compared with $3.02$
for F-blind, giving $\Delta(\mathrm{ttr}) = +4.29 \pm 1.77$. The advantage all but vanishes when
refreshes only reveal the state, where $\Delta(\mathrm{ttr})$ reaches at most $+0.14$, about a
thirtieth of the causal effect on the same scale. Within the causal environment the contrast
grows from an uncertain $+1.07 \pm 3.62$ at periodic to the large irregular-schedule effect,
matching the predicted dose-response. The ladder is not unresponsive at periodic: its rungs still
span $4.07$ to $6.05$ there, but ttr duplicates what the step count already gives. The advantage
also fails to appear when a sensor-faithful update announces that the actuator has latched; there
it reverses ($-0.75 \pm 0.73$), with A-ZOH climbing from $1.36$ to $8.36$ and overtaking F's
$7.67$. Under reveal, where the target is zero, no rung exceeds $0.48$, so the causal gap
reflects discrimination rather than reactivity to a changed input.

\input{sections/table_toy}

This act makes the preceding nulls interpretable. The same ladder, the same contrast and the same
evaluation procedure detect a large effect exactly when refresh timing changes the plant and
cannot be inferred from another input. Across the three acts, schedule knowledge is therefore useful in proportion to how
strongly the schedule participates in the dynamics, rather than simply because future refresh
times are known.

%% file: sections/table_main.tex
\begin{table}[H]
\centering
\footnotesize
\captionsetup{font=small, skip=4pt}
\setlength{\tabcolsep}{4.5pt}
\renewcommand{\arraystretch}{1.00}

\begin{adjustbox}{max width=\textwidth,center}
\begin{tabular}{llllcccc}
\toprule
\textbf{Study} &
\textbf{Protocol} &
\textbf{Environment} &
\textbf{Schedule} &
\textbf{A-ZOH} &
\textbf{F-blind} &
\textbf{F} &
\boldmath$\Delta(\mathrm{ttr})$ \\
\midrule

% ============================================================
% Prediction
% ============================================================

\multirow{20}{*}{\textbf{Prediction}}
& \multirow{10}{*}{Cut-off}
& \multirow{3}{*}{Ant}
    & periodic & $+0.8240$ & $+0.8003$ & $+0.7930$ & $-0.0073 \pm 0.0220$ \\
&
&
    & $\pm 10\%$ & $+0.8095$ & $+0.8048$ & $+0.7938$ & $-0.0110 \pm 0.0382$ \\
&
&
    & $\pm 50\%$ & $+0.8095$ & $+0.8482$ & $+0.8387$ & $-0.0095 \pm 0.0671$ \\

\cmidrule(lr){3-8}

&
&
\multirow{3}{*}{HalfCheetah}
    & periodic & $+0.8260$ & $+0.8504$ & $+0.8064$ & $-0.0440 \pm 0.1411$ \\
&
&
    & $\pm 10\%$ & $+0.7653$ & $+0.7950$ & $+0.8068$ & $+0.0119 \pm 0.1845$ \\
&
&
    & $\pm 50\%$ & $+0.8430$ & $+0.8504$ & $+0.8556$ & $+0.0053 \pm 0.0951$ \\

\cmidrule(lr){3-8}

&
&
\multirow{3}{*}{Walker2d}
    & periodic & $+0.8512$ & $+0.7118$ & $+0.8287$ & $+0.1169 \pm 0.8443$ \\
&
&
    & $\pm 10\%$ & $+0.8094$ & $+0.8316$ & $+0.8938$ & $+0.0622 \pm 0.2267$ \\
&
&
    & $\pm 50\%$ & $+0.7976$ & $+0.7276$ & $+0.8817$ & $+0.1542 \pm 0.2698$ \\

&
&
\multicolumn{2}{r}{\emph{pooled} ($n{=}27$)}
    & & & & $+0.0310 \pm 0.0490$ \\

\cmidrule(lr){2-8}

&
\multirow{10}{*}{Sensor-faithful}
& \multirow{3}{*}{Ant}
    & periodic & $+0.8657$ & $+0.8603$ & $+0.8541$ & $-0.0062 \pm 0.0370$ \\
&
&
    & $\pm 10\%$ & $+0.8526$ & $+0.8684$ & $+0.8610$ & $-0.0073 \pm 0.1145$ \\
&
&
    & $\pm 50\%$ & $+0.8895$ & $+0.9112$ & $+0.9247$ & $+0.0135 \pm 0.0521$ \\

\cmidrule(lr){3-8}

&
&
\multirow{3}{*}{HalfCheetah}
    & periodic & $+0.9917$ & $+0.9870$ & $+0.9923$ & $+0.0054 \pm 0.0150$ \\
&
&
    & $\pm 10\%$ & $+0.9953$ & $+0.9942$ & $+0.9943$ & $+0.0001 \pm 0.0026$ \\
&
&
    & $\pm 50\%$ & $+0.9969$ & $+0.9935$ & $+0.9955$ & $+0.0020 \pm 0.0059$ \\

\cmidrule(lr){3-8}

&
&
\multirow{3}{*}{Walker2d}
    & periodic & $+0.9768$ & $+0.9469$ & $+0.9349$ & $-0.0120 \pm 0.0232$ \\
&
&
    & $\pm 10\%$ & $+0.9900$ & $+0.9848$ & $+0.9835$ & $-0.0013 \pm 0.0008$ \\
&
&
    & $\pm 50\%$ & $+0.9912$ & $+0.9846$ & $+0.9864$ & $+0.0018 \pm 0.0034$ \\

&
&
\multicolumn{2}{r}{\emph{pooled} ($n{=}27$)}
    & & & & $-0.0005 \pm 0.0066$ \\

\midrule

% ============================================================
% Control
% ============================================================

\multirow{6}{*}{\textbf{Control}}
& \multirow{6}{*}{Fixed CEM}
& \multirow{2}{*}{Ant}
    & periodic & $349.9$ & $377.5$ & $408.6$ & $+31 \pm 102$ \\
&
&
    & $\pm 50\%$ & $254.5$ & $373.5$ & $300.5$ & $-73 \pm 44$ \\

\cmidrule(lr){3-8}

&
&
\multirow{2}{*}{HalfCheetah}
    & periodic & $78.4$ & $38.3$ & $60.7$ & $+22 \pm 44$ \\
&
&
    & $\pm 50\%$ & $165.1$ & $167.5$ & $142.3$ & $-25 \pm 50$ \\

\cmidrule(lr){3-8}

&
&
\multirow{2}{*}{Walker2d}
    & periodic & $18.3$ & $47.6$ & $107.5$ & $+60 \pm 390$ \\
&
&
    & $\pm 50\%$ & $140.0$ & $216.2$ & $93.6$ & $-123 \pm 356$ \\

\bottomrule
\end{tabular}
\end{adjustbox}

\caption{
\textbf{Prediction and control.} Prediction means and 95\% $t$-intervals pair three training seeds
within each cell and all $27$ differences on pooled rows. Control means and intervals pair
evaluation seeds --- six for Ant and three otherwise.
Prediction reports slow-tier ZOH skill at $H{=}50$; control reports CEM return.
}
\label{tab:prediction_control}

\end{table}

%% file: sections/table_toy.tex
% ============================================================
% TABLE 2: TOY PROBE
% ============================================================

\begin{table}[H]
\centering
\footnotesize
\captionsetup{font=small, skip=8pt}
\setlength{\tabcolsep}{4.5pt}
\renewcommand{\arraystretch}{1.08}

\begin{adjustbox}{max width=\textwidth,center}
\begin{tabular}{llllcccc}
\toprule
\textbf{Study} &
\textbf{Protocol} &
\textbf{World} &
\textbf{Schedule} &
\textbf{A-ZOH} &
\textbf{F-blind} &
\textbf{F} &
\boldmath$\Delta(\mathrm{ttr})$ \\
\midrule

\multirow{8}{*}{\textbf{Paired-world probe}}
& \multirow{4}{*}{Cut-off}
& \multirow{2}{*}{Reveal}
    & periodic & $0.29$ & $0.34$ & $0.48$ & $+0.14 \pm 0.13$ \\
&
&
    & $\pm 50\%$ & $0.17$ & $0.19$ & $0.20$ & $+0.02 \pm 0.17$ \\

\cmidrule(lr){3-8}

&
&
\multirow{2}{*}{Causal}
    & periodic & $4.07$ & $4.98$ & $6.05$ & $+1.07 \pm 3.62$ \\
&
&
    & $\pm 50\%$ & $1.36$ & $3.02$ & $7.31$ & $+4.29 \pm 1.77$ \\

\cmidrule(lr){2-8}

&
\multirow{4}{*}{Sensor-faithful}
& \multirow{2}{*}{Reveal}
    & periodic & $0.29$ & $0.34$ & $0.48$ & $+0.14 \pm 0.07$ \\
&
&
    & $\pm 50\%$ & $0.17$ & $0.18$ & $0.22$ & $+0.04 \pm 0.22$ \\

\cmidrule(lr){3-8}

&
&
\multirow{2}{*}{Causal}
    & periodic & $7.52$ & $7.66$ & $7.52$ & $-0.14 \pm 2.06$ \\
&
&
    & $\pm 50\%$ & $8.36$ & $8.42$ & $7.67$ & $-0.75 \pm 0.73$ \\

\bottomrule
\end{tabular}
\end{adjustbox}

\caption{
\textbf{Paired-world probe.} Two worlds share a command stream and differ only in when the
next slow channel refreshes, at step $37$ or $44$, so ttr alone distinguishes them until
step $37$. Entries are the peak separation of a rung's two predicted velocity
trajectories, in raw velocity units, as three-seed means; $\Delta(\mathrm{ttr})$ is paired
within cell with a 95\% $t$-interval. The target is set by the worlds, not by any model:
$9.14$ under \textbf{causal}, and exactly $0$ under \textbf{reveal}, where any separation
is spurious.
}
\label{tab:toy_probe}

\end{table}

%% file: sections/related_work.tex
\section{Related Work}

World models learn dynamics for prediction, planning, and policy optimization, spanning recurrent
state-space models \citep{hafner2020dreamer,hafner2023dreamerv3,vishe2026skill,shan2026cyclegen}, discrete-token
models such as IRIS \citep{micheli2023iris}, diffusion models such as DIAMOND
\citep{alonso2024diamond}, and latent planners such as TD-MPC2 \citep{hansen2024tdmpc2}. Despite
substantial architectural progress and deployment on physical systems \citep{wu2023daydreamer},
this literature generally assumes that a complete observation arrives at every environment step. Related multimodal reasoning and evaluation settings increasingly probe models across visual, and auditory observations \citep{mundada2025wildscore, cysmr, vishe2025musecpbench}
Work on irregular time series relaxes uniform sampling through continuous-time dynamics
\citep{chen2018neuralode,rubanova2019latentode,kidger2020neuralcde}, time-aware attention
\citep{shukla2021mtan,chen2024contiformer}, and explicit missingness features. GRU-D
\citep{che2018grud}, in particular, encodes elapsed time since the previous observation,
corresponding closely to our staleness feature. These methods model when past observations
occurred; we additionally isolate the value of knowing when the next observation will arrive.

Zero-order holds, intermittent observations, and event-triggered sampling are established topics
in sampled-data and networked control \citep{sinopoli2004kalman,heemels2012event}, where staleness
is closely related to age of information \citep{kaul2012aoi,yates2021aoisurvey}. Delayed and
concurrent reinforcement learning similarly studies systems in which sensing or actuation does not
occur immediately \citep{ramstedt2019realtime,xiao2020thinking,bouteiller2021randomdelays}. Prior
work therefore establishes that timing can matter, but does not isolate the forward-looking portion
of a known schedule as an input to an action-conditioned learned world model. Our study makes that
comparison directly and distinguishes schedules that merely determine when state is observed from
schedules whose refresh events participate in the dynamics.

%% file: sections/conclusion.tex
\section{Conclusion}

In this work, we studied the role of sampling schedule information in world models operating under mixed-frequency observations. Our results show that when refreshes only provide new observations, time-to-refresh does not improve prediction across different environments, levels of schedule irregularity, and evaluation settings. When refresh events directly affect the system dynamics, however, time-to-refresh becomes informative, with larger improvements as the sampling schedule becomes more irregular. These findings provide a practical distinction for incorporating schedule information into world models. Staleness provides useful information about previously observed measurements, while forward-looking refresh timing becomes useful when refresh events participate in the system dynamics. This suggests that the choice of schedule information should depend on whether a refresh only reveals the system state or directly affects how that state evolves.

\section{Limitations}
Our evaluation is limited to state-based observations and a fixed family of refresh schedules.
Future work should consider multimodal world models, communication delays, and larger physical
systems. \textbf{LLM usage disclosure:} AI tools were used to assist in creating illustrative figures.

%% file: sections/appendix.tex
\section{Experimental Detail}
\label{app:setup}

Tables~\ref{tab:envs} and~\ref{tab:hyperparams} give the environment statistics and the
architecture and optimization settings, which are identical across every rung and environment.
The planner of \S\ref{sec:control} evaluates 256 candidate action sequences per decision, retains
50 elites over five refinement iterations, aggregates predictions pessimistically using the
minimum over the three seed checkpoints, and represents candidate actions as bounded residuals
around the trained SAC policy. Planning uses horizon 12 for Ant and Walker2d and 25 for
HalfCheetah. Each evaluation seed averages five episodes of up to 500 steps.
Table~\ref{tab:prediction_control} aggregates six evaluation seeds for Ant and three for
HalfCheetah and Walker2d. The reference planner replaces the learned dynamics with full-rate
oracle dynamics in the same controller. Its mean return over three evaluation seeds is $440$,
$936$ and $769$ on Ant, HalfCheetah and Walker2d, respectively.

The prediction grid is the full crossing of three rungs, three environments, three schedules and
three seeds, giving \textbf{81 training runs}. Constructed-environment datasets are generated per
\texttt{(variant, schedule, seed)} for the coupling reason given in \S\ref{sec:data}. The
closed-loop study crosses the same rungs and environments with the periodic and $\pm50\%$
schedules. Each cell is evaluated over six seeds on Ant and three seeds on HalfCheetah and
Walker2d. The refresh-gated study crosses three rungs, two worlds, two schedules and three seeds,
giving \textbf{36 training runs}.

Prediction contrasts pair F against F-blind within a cell --- the same environment, schedule and
training seed --- so that each difference holds constant everything but the ttr column. A
prediction per-cell interval is a two-sided 95\% $t$-interval over three paired training-seed
differences; a pooled prediction interval uses all $27$ paired differences from that protocol.
Control contrasts are instead paired by evaluation seed, giving six paired differences for Ant
and three for HalfCheetah and Walker2d.

\begin{table}[h]
\centering\small
\caption{$\Delta(\mathrm{ttr})$ in slow-tier ZOH skill against rollout horizon, for the two
endpoint schedules. Each cell pools the nine paired differences of that protocol and schedule
(three environments $\times$ three seeds) with a 95\% $t$-interval. Under sensor-faithful the
contrast shrinks with horizon; under cut-off it is non-monotone, rising to $H{=}15$ and falling
back by $H{=}50$. One cell of the twelve excludes zero, cut-off at $\pm$50\% and $H{=}15$, at an
intermediate horizon under the protocol in which no real observation ever arrives; it does not
persist to the horizon we report. Growth to the longest horizon is the signature a horizon-scoped
feature would leave, and neither protocol shows it.}
\label{tab:horizon}
\begin{tabular}{llccc}
\toprule
\textbf{Protocol} & \textbf{Schedule} & $H{=}5$ & $H{=}15$ & $H{=}50$ \\
\midrule
Cut-off & periodic & $+0.0022 \pm 0.0201$ & $+0.0203 \pm 0.0324$ & $+0.0219 \pm 0.1439$ \\
Cut-off & $\pm$50\% & $+0.0239 \pm 0.0434$ & $+0.0576 \pm 0.0385$ & $+0.0500 \pm 0.0755$ \\
Sensor-faithful & periodic & $+0.0036 \pm 0.0101$ & $+0.0033 \pm 0.0197$ & $-0.0043 \pm 0.0093$ \\
Sensor-faithful & $\pm$50\% & $+0.0173 \pm 0.0224$ & $+0.0107 \pm 0.0156$ & $+0.0057 \pm 0.0093$ \\
\bottomrule
\end{tabular}
\end{table}

\subsection{Aligning the training objective to the evaluation horizon}
\label{app:rollout}

Models are trained on one-step prediction but scored on a 50-step rollout, so a horizon-scoped
feature such as ttr receives little gradient and the Act~1 null could be an artefact of that
mismatch rather than of the setting. We therefore fine-tuned the Act~1 checkpoints from the
periodic and $\pm50\%$ cells with the loss computed on the rollout itself, under the
sensor-faithful protocol, and compared them against teacher-forced continuations matched on base
checkpoint, epochs, learning rate and selection metric, so that the contrast is the objective
rather than the additional training. Ordinary training is unchanged by this and every number
elsewhere in the paper comes from the original one-step runs. The grid covers the three rungs,
three environments, the periodic and $\pm50\%$ schedules and three seeds in both arms.

Two quantities matter. The \emph{gate}, the rollout-aligned arm minus the teacher-forced control,
asks whether the intervention did anything at all; if it is zero, the arms are indistinguishable
and no conclusion about ttr can be drawn from them. Pooled over the grid, the gate is
$+0.0043 \pm 0.0020$ and excludes zero, and it excludes zero in five of the six
metric-by-environment readings in Table~\ref{tab:rollout}. Only HalfCheetah's skill column fails
it, which is expected where skill already sits above $0.98$ and has no headroom left to move.

With the gate passed, $\Delta(\mathrm{ttr})$ is still zero. Pooled it moves from
$+0.0007 \pm 0.0064$ before fine-tuning to $-0.0043 \pm 0.0051$ after, so aligning the objective
to the horizon does not reveal an effect. One cell of the twelve excludes zero, and it is negative, meaning ttr cost a little there; none is positive. The null is
therefore not a consequence of training one step ahead.

\begin{table}[h]
\centering\small
\caption{Fine-tuning with the objective aligned to the 50-step evaluation rollout. The gate is the
rollout-aligned arm minus its matched teacher-forced control, so a non-zero value means the
intervention took effect; $\Delta(\mathrm{ttr})$ is then measured within the rollout-aligned arm.
Both metrics are evaluated on the slow tier: ZOH skill is defined in \S\ref{sec:eval}, and
transition MSE restricts squared prediction error to steps where the absolute difference between
the ground truth and the held value exceeds $0.25$ times that channel's standard deviation.
Entries are means with 95\% $t$-intervals, and $^{*}$ marks an interval excluding zero. Positive
always means ttr helped, which for the error metric means the difference is taken the other way
round.}
\label{tab:rollout}
\input{tables/t13_act3_3env}
\end{table}

\begin{table}[h]
\centering\small
\caption{Environments and trained-policy datasets. The slow tier is root linear and angular
velocity in all three; the policy is SAC trained for 300K steps per environment.}
\label{tab:envs}
\begin{tabular}{lccccc}
\toprule
\textbf{Environment} & \textbf{obs\_dim} & \textbf{act\_dim} & \textbf{episodes} & \textbf{observations} & \textbf{slow-tier channels} \\
\midrule
Ant-v4         & 27 & 8 & 400 & 381K & 6 \\
HalfCheetah-v4 & 17 & 6 & 400 & 400K & 3 \\
Walker2d-v4    & 17 & 6 & 400 & 319K & 3 \\
\bottomrule
\end{tabular}
\end{table}

\begin{table}[h]
\centering\small
\caption{Architecture and optimization, identical across every rung and environment.}
\label{tab:hyperparams}
\begin{tabular}{llll}
\toprule
\textbf{Parameter} & \textbf{Value} & \textbf{Parameter} & \textbf{Value} \\
\midrule
$d_{\mathrm{model}}$   & 128   & Optimizer        & AdamW \\
Transformer layers   & 4     & Learning rate    & $3 \times 10^{-4}$ \\
Attention heads      & 4     & Weight decay     & $10^{-2}$ \\
Context length       & 32    & Gradient clip    & 1.0 \\
Batch size           & 128   & LR schedule      & Cosine annealing \\
Epochs               & 15    & Train/val/test   & 80\,/\,10\,/\,10 by episode \\
\bottomrule
\end{tabular}
\end{table}

\section{Detailed Analysis}
\label{app:analysis}

Tables~\ref{tab:prediction_control} and~\ref{tab:toy_probe} report the studies: the three rungs
as measured, then the contrast derived from them. Read as a whole they say one thing. Six of the
thirty-two cells have an interval excluding zero, but only one is both positive and large: the
constructed environment where a refresh latches the actuator, on the irregular schedule, scored
without further sensor data, at $+4.29$ on a scale whose target separation is $9.14$. Three of the
other five are negative, meaning ttr cost a little; the remaining two are the \textbf{reveal}
cells at $+0.14$ on that same scale, some thirty times smaller than the causal effect. The rest of this appendix
reads that pattern along the three axes that vary the refresh event itself: whether it changes the
dynamics (\textbf{causal} against \textbf{reveal}), whether its timing is inferable unaided
($\pm$50\% against periodic), and whether it announces itself in the data (cut-off against
sensor-faithful).

\subsection{When Refresh Timing Pays}
\label{sec:conditions}

Each isolates one condition, and the effect survives only where all three hold.

\textbf{The refresh must cause rather than report}, as a held command latched on a bus tick does
and a fixed-rate IMU or GPS does not. Under \textbf{causal} at $\pm$50\% and cut-off, F separates
the two worlds by $7.31$ of the true $9.14$ against F-blind's $3.02$ ($+4.29 \pm 1.77$). Under
\textbf{reveal} the dynamics never consult the schedule, the worlds are bit-identical, and no rung
separates them by more than $0.48$ --- and since each receives the same changed ttr there, the
causal gap is discrimination between worlds that differ, not reactivity to an input that changed.

\textbf{The timing must not be inferable another way.} A fixed-rate bus is a matter of counting,
and so is a periodic schedule: ttr duplicates the step index even though the refresh is fully
causal, and the contrast falls to $+1.07 \pm 3.62$. The rungs still span $4.07$ to $6.05$ there,
so this is ttr becoming redundant rather than the environment going quiet --- and a genuine effect
must therefore \emph{grow} from periodic to $\pm$50\%, which it does.

\textbf{The event must not announce itself through the data}, as it does on a shared-clock node
that reads and actuates in one slot. A refresh delivering the real observation already reveals
that the latch happened, and under sensor-faithful the advantage duly disappears
($-0.75 \pm 0.73$), A-ZOH climbing from $1.36$ to $8.36$ and overtaking F's $7.67$.

\textbf{Prediction fails the first condition outright.} The MuJoCo refresh mask is drawn from
\texttt{(seed, episode index)} and the physics never reads it, so no amount of jitter can make ttr
informative. Pooled over all 27 cells $\Delta(\mathrm{ttr})$ is $+0.0310 \pm 0.0490$ under cut-off
and $-0.0005 \pm 0.0066$ under sensor-faithful, every per-environment interval contains zero, and
the effect does not grow with jitter. The second estimate is near zero with a tight 95\%
confidence interval, from $-0.0071$ to $+0.0061$, rather than merely being an isolated
non-significant cell. Separately, the Act~3 result shows that the model and paired-world probe can
detect a ttr effect where the condition is met, producing $+4.29$ on the raw trajectory-separation
metric. Three checks close the obvious objections: the effect does not grow with rollout horizon
(Table~\ref{tab:horizon}); it survives
fine-tuning the same checkpoints with the objective computed on the 50-step rollout, against a
matched teacher-forced control (\S\ref{app:rollout}); and it is not a ceiling artefact, since the cut-off column is far from saturation while
the sensor-faithful one is not --- HalfCheetah sits at or above $0.987$ on every sensor-faithful
cell --- and both give the same null.

\textbf{Control satisfies it only weakly.} Observations drive actions and actions drive the state,
but by a long path and never as a latch --- the planner's action applies at every step, exactly as
under \textbf{reveal}. The claim permits a small effect; the rung differences are real but small,
inconsistent in sign, and dwarfed by the distance to a reference planner given fresh observations
on every channel, which returns $440$, $936$ and $769$ on Ant, HalfCheetah and Walker2d. Ant's
best rung reaches $408.6$ of its $440$, while HalfCheetah and Walker2d reach $167.5$ and $216.2$.
Fidelity, not schedule knowledge, is the binding constraint. It also bounds the claim: the planner
recovers most of that reference only on Ant (85--93\%), where $\Delta(\mathrm{ttr})$ contains zero
at periodic and is \emph{negative} at $\pm$50\%, and falls far short on HalfCheetah (8--18\%) and
Walker2d (14--28\%), where rung differences are differences between failure modes ($\pm390$ and
$\pm356$). We report the
closed-loop test as the single-environment result it is.

Taken together the three conditions describe an irregularly-clocked held actuator observed over a
separate channel --- a property of a system's wiring, checkable before any model is trained. A
sensor schedule and an actuation schedule are different objects: telling a world model when its
sensors will report is wasted effort, because those times do not enter the dynamics, while telling
it when its actuators will update is not, provided the update is genuinely gated and its timing is
not already implicit in the data.

%% file: tables/t13_act3_3env.tex
% GENERATED by scripts/emit_threeenv_tables.py -- do not edit by hand.
% Act III across three envs and two metrics
\begin{tabular}{llccc}
\toprule
metric & env & gate (aligned $-$ teacher-forced) & $\Delta$(ttr) @ $\pm$50\% & $\Delta$(ttr) @ periodic \\
\midrule
ZOH skill & Ant & +0.0083 {\scriptsize $\pm$ 0.0054}$^{*}$ & +0.0035 {\scriptsize $\pm$ 0.0328} & -0.0159 {\scriptsize $\pm$ 0.0401} \\
 & HalfCheetah & +0.0007 {\scriptsize $\pm$ 0.0010} {\scriptsize\emph{(gate fails)}} & -0.0003 {\scriptsize $\pm$ 0.0020} & -0.0007 {\scriptsize $\pm$ 0.0028} \\
 & Walker2d & +0.0039 {\scriptsize $\pm$ 0.0026}$^{*}$ & -0.0007 {\scriptsize $\pm$ 0.0035} & -0.0115 {\scriptsize $\pm$ 0.0155} \\
\addlinespace
transition MSE & Ant & +0.1343 {\scriptsize $\pm$ 0.0890}$^{*}$ & +0.0357 {\scriptsize $\pm$ 0.3301} & -0.2075 {\scriptsize $\pm$ 0.3710} \\
 & HalfCheetah & +0.0013 {\scriptsize $\pm$ 0.0011}$^{*}$ & -0.0014 {\scriptsize $\pm$ 0.0060} & -0.0008 {\scriptsize $\pm$ 0.0036} \\
 & Walker2d & +0.0072 {\scriptsize $\pm$ 0.0048}$^{*}$ & -0.0010 {\scriptsize $\pm$ 0.0056} & -0.0175 {\scriptsize $\pm$ 0.0162}$^{*}$ \\
\bottomrule
\end{tabular}